%% file: neurips_2025.tex
\documentclass{article}

 \usepackage[main, final]{neurips_2025}

\usepackage[utf8]{inputenc} 
\usepackage[T1]{fontenc}    
\usepackage{hyperref}       
\usepackage{url}            
\usepackage{booktabs}       
\usepackage{amsfonts}       
\usepackage{nicefrac}       
\usepackage{microtype}      
\usepackage{xcolor}         
\usepackage{amsmath}
\usepackage{cleveref}
\usepackage{graphicx}
\usepackage{caption}   
\usepackage{booktabs}
\usepackage{multirow}
\usepackage{makecell}

\title{Uncovering Graph Reasoning in Decoder-only Transformers with Circuit Tracing}

\author{%
  Xinnan Dai$^1$\thanks{Accepted by Workshop on
Efficient Reasoning. Corresponding: guokai1@msu.edu} \
  Chung-Hsiang Lo$^2$\
  Kai Guo$^1$\
  Shenglai Zeng$^1$\
  Dongsheng Luo$^3$\
  Jiliang Tang$^1$\
\\
    $^1$Michigan State University, \
  $^2$Northeastern University, \
$^3$Florida International University \\
}

\begin{document}

\maketitle

\begin{abstract}

Transformer-based LLMs demonstrate strong performance on graph reasoning tasks, yet their internal mechanisms remain underexplored. To uncover these reasoning process mechanisms in a fundamental and unified view, we set the basic decoder-only transformers and explain them using the circuit-tracer framework. Through this lens, we visualize reasoning traces and identify two core mechanisms in graph reasoning: token merging 
and structural memorization 
, which underlie both path reasoning and substructure extraction tasks. We further quantify these behaviors and analyze how they are influenced by graph density and model size. Our study provides a unified interpretability framework for understanding structural reasoning in decoder-only Transformers.

\end{abstract}

\input{section/01_intro}

\input{section/03_Experiments}

\input{section/02_Related_work_and_Background}

\section{Conclusion}
\vspace{-0.1cm}
We provide a unified perspective on how decoder-only Transformers solve graph reasoning tasks, revealing token merging and structure memorization as core mechanisms. These behaviors vary with task complexity and model scale, offering a compact framework for interpreting structural reasoning in Transformers.

\bibliographystyle{unsrt}
\bibliography{references}








\appendix

\section{Appendices in Experiments}
\label{app:exp}
In this section, we provide additional details of our experimental setup. The graph datasets vary across tasks, and for each task, the graph configurations, Transformer architectures, and transcoder training procedures are customized accordingly. 
We adopt the GPT-2 model with ROPE positional embeddings as the backbone. While transformers are capable of achieving high accuracy, we employ a lightweight variant of the GPT-2 architecture consisting of only 5 layers. After training the transformers on the specified tasks, we train cross-layer transcoders at each layer. These transcoders are then merged to construct an attribution graph according to~\citep{ameisen2025circuit}, which is used to reveal the implicit internal structures of the transformer for interpretability. 
Additionally, we report the parameters used in our visualizations. The summary of these experimental settings is presented in Table~\ref{tab:exp_set}. All of the experiments are run on a single RTX A6000.

\section{Structure memorization in training}
\label{app:mem_train}
We find that Transformers tend to memorize structural patterns during pretraining. To investigate this, we construct training data by extracting subgraphs from backbone graphs with varying densities (0.2 and 0.4). Each sampled subgraph contains at most 10 nodes, and we focus on path reasoning tasks to evaluate whether Transformers can memorize edge combinations during training.
We evaluate accuracy on the test set using three criteria: 1. Local Accuracy: The predicted path's edges are fully contained within the given subgraph; 2. Exist Accuracy: The predicted path's edges exist somewhere in the original backbone graph; 3. Global Accuracy: Given only a start node and an end node, the model must predict a correct path in the backbone graph without explicit subgraph context. The maximum path length is also recorded to assess the length of the predicted path.

\begin{figure}[htbp]
  \centering

  \begin{minipage}[c]{0.45\textwidth}
   \centering
    
    \includegraphics[width=\linewidth]{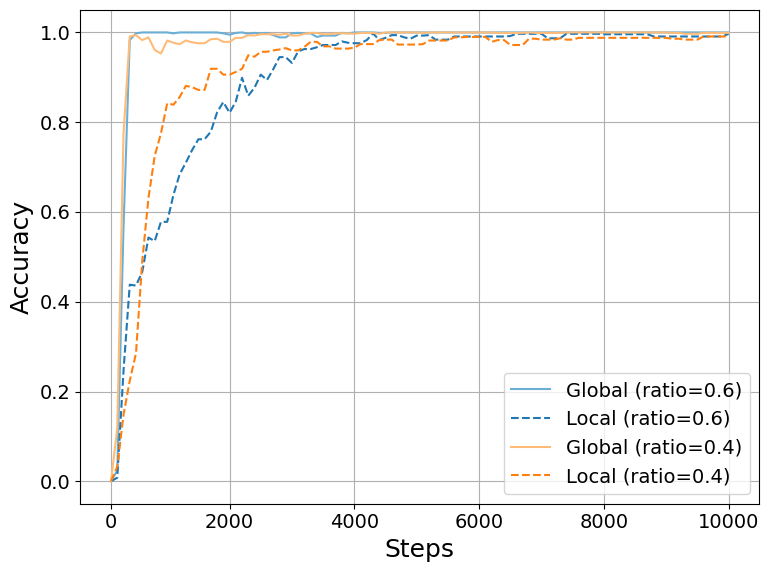}
    \caption{Local Acc and Global Acc change with training steps}
    \label{fig:hallucination}

  \end{minipage}
    \hfill
  \begin{minipage}[c]{0.45\textwidth}
    \centering
    \includegraphics[width=\linewidth]{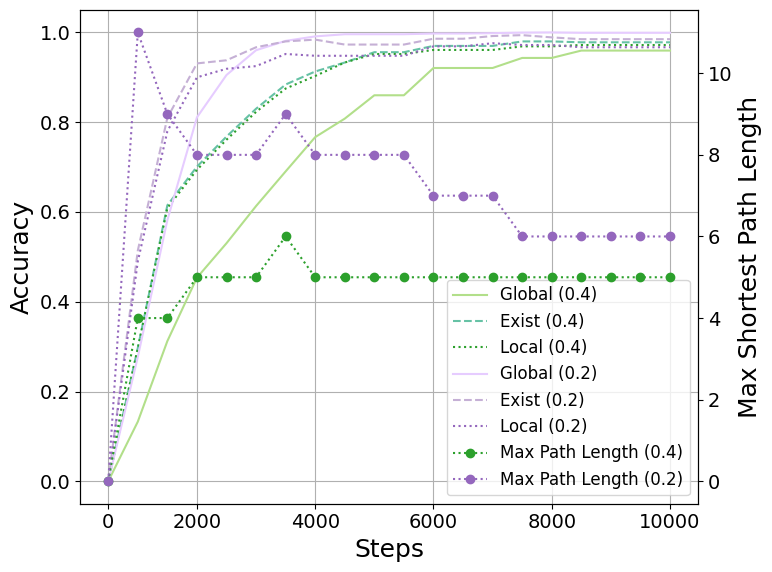} \\
    \caption{Local Acc, Exist Acc, Global Acc, and Max Path Length change with training steps}
    \label{fig:global_figure}
  \end{minipage}
  
\end{figure}

First, we evaluate the case where the backbone graph contains 10 nodes, and we randomly drop 40\% or 60\% of its edges. We measure both Global Accuracy and Local Accuracy during training, shown in Figure~\ref{fig:hallucination}. We observe that Transformers initially memorize the full backbone graph, reaching 100\% Global Accuracy at early training stages. In contrast, Local Accuracy improves more gradually, suggesting that Transformers first memorize the global structure before adapting to the local subgraphs. Additionally, a lower edge dropout ratio leads to faster learning, indicating that denser graphs facilitate structural memorization.

Next, we increase the backbone graph size to 50 nodes and evaluate the setting where subgraphs are sampled by selecting nodes only—without dropping any edges among the selected nodes. The results are presented in Figure~\ref{fig:global_figure}. As the graph size increases, memorizing the full backbone structure becomes more challenging, and we observe that Global Accuracy improves more slowly compared to the 10-node case, while Local Accuracy increases more quickly. Interestingly, Exist Accuracy exceeds Local Accuracy, suggesting that the model continues to rely on memorized global edges even in local contexts. Moreover, despite training only on short paths (3–5 hops), the model is capable of predicting accurate paths exceeding 10 hops, implying that Transformers may prioritize high-probability paths rather than strictly the shortest ones—even when trained under shortest-path supervision.

\input{table/parameters}

\section{Additional Figures}
\label{app:figs}
\begin{figure}[htbp]
  \centering

  \begin{minipage}[c]{0.55\textwidth}
    \centering
    
    \includegraphics[width=\linewidth]{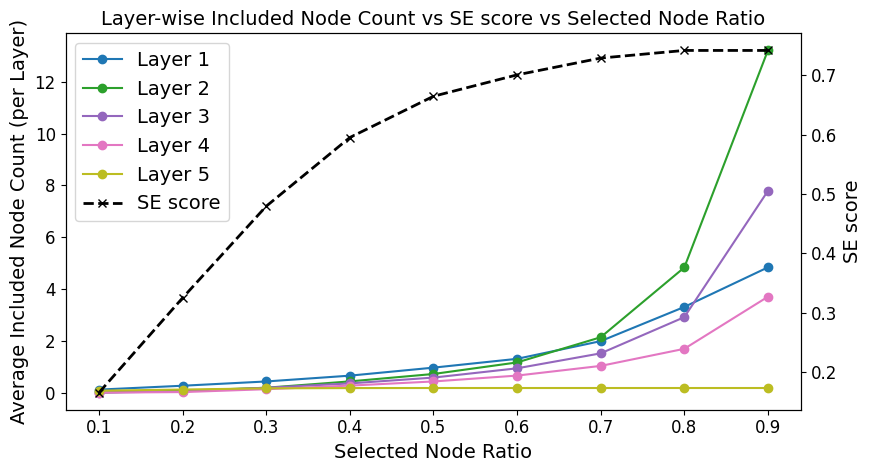}
    \caption{Pattern learned across layers. The counting demonstrates the number of tokens merged across the layers. The token merging never shows in layer 4 and layer Output.}
    \label{fig:substructure}
  \end{minipage}%
  \hfill
    \begin{minipage}[c]{0.35\textwidth}
      \centering
    \captionof{table}{Token merged in attributed graph reasoning. $\mathrm{S_E}$ is 0.94 with the selected node ratio of 0.9. Start, End, and Edge denote attributes merged from the start node, end node, and full edge, respectively.}
    \label{tab:attr_graph}
\resizebox{\linewidth}{!}{
  \begin{tabular}{llll}
  \toprule
    & \textbf{Start} & \textbf{End} & \textbf{Edge} \\
    \midrule
    L1        & 0     & 6     & 3   \\
    L2        & 39    & 40    & 39  \\
    L3        & 225   & 218   & 222 \\
    L4-Out    & 0     & 0     & 0   \\
    Overall   & 264   & 264   & 264 \\
    \bottomrule
  \end{tabular}
}
  \end{minipage}%
\end{figure}

\begin{figure}[htbp]
\vspace{-0.6cm}
  \centering

  \begin{minipage}[c]{0.7\textwidth}
   \centering
    
    \includegraphics[width=\linewidth]{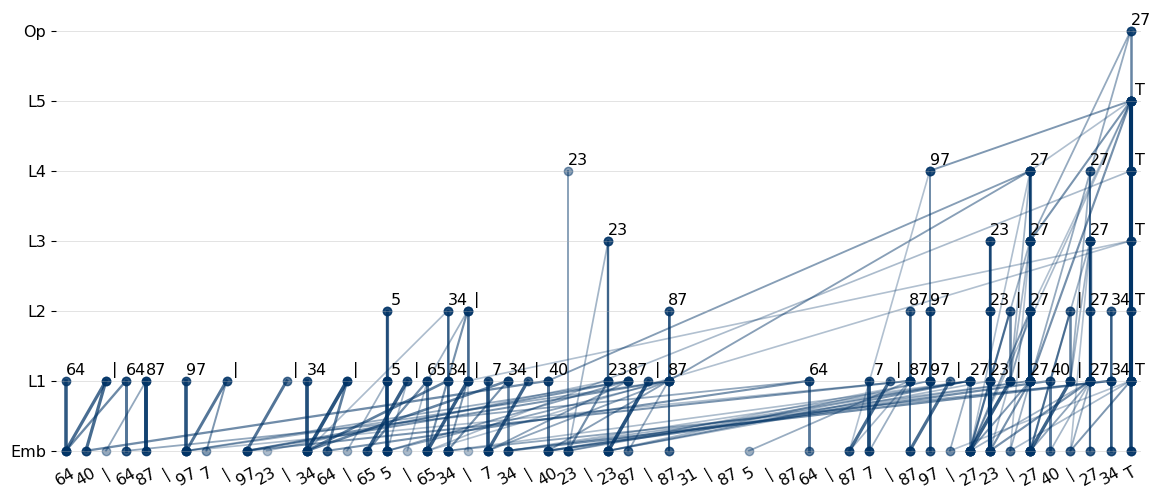}
    \caption{Circuit tracer in the pattern extraction task. The predicted patterns are (27, 23, 40) and (97, 87, 7).}
    \label{fig:patt_vis}
  \end{minipage}
\hfill
  \hfill
  \begin{minipage}[c]{0.25\textwidth}
    \centering
    \includegraphics[width=\linewidth]{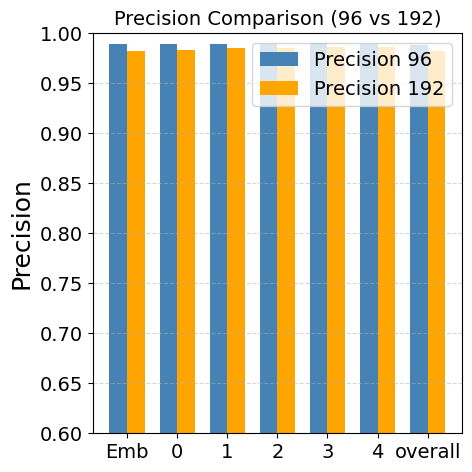} \\
    \caption{Precision under different hidden dimensions}
    \label{fig:pre}    \label{fig:mem_vis}
  \end{minipage}
  \vspace{-0.6cm}
\end{figure}

\section{Limitation}
Our analysis is limited to decoder-only Transformers trained from scratch, and a gap remains between these models and real-world large language models. Additionally, a more thorough investigation into the effects and limitations of circuit tracing should be included in future work.
\section{Boarder Impact}
We uncover the core mechanisms behind how decoder-only Transformers perform graph reasoning. Given that graph reasoning is a specialized form of structured reasoning, these insights may extend to general reasoning tasks in large language models



\newpage
\section*{NeurIPS Paper Checklist}

The checklist is designed to encourage best practices for responsible machine learning research, addressing issues of reproducibility, transparency, research ethics, and societal impact. Do not remove the checklist: {\bf The papers not including the checklist will be desk rejected.} The checklist should follow the references and follow the (optional) supplemental material.  The checklist does NOT count towards the page
limit. 

Please read the checklist guidelines carefully for information on how to answer these questions. For each question in the checklist:
\begin{itemize}
    \item You should answer \answerYes{}, \answerNo{}, or \answerNA{}.
    \item \answerNA{} means either that the question is Not Applicable for that particular paper or the relevant information is Not Available.
    \item Please provide a short (1–2 sentence) justification right after your answer (even for NA). 
\end{itemize}

{\bf The checklist answers are an integral part of your paper submission.} They are visible to the reviewers, area chairs, senior area chairs, and ethics reviewers. You will be asked to also include it (after eventual revisions) with the final version of your paper, and its final version will be published with the paper.

The reviewers of your paper will be asked to use the checklist as one of the factors in their evaluation. While "\answerYes{}" is generally preferable to "\answerNo{}", it is perfectly acceptable to answer "\answerNo{}" provided a proper justification is given (e.g., "error bars are not reported because it would be too computationally expensive" or "we were unable to find the license for the dataset we used"). In general, answering "\answerNo{}" or "\answerNA{}" is not grounds for rejection. While the questions are phrased in a binary way, we acknowledge that the true answer is often more nuanced, so please just use your best judgment and write a justification to elaborate. All supporting evidence can appear either in the main paper or the supplemental material, provided in appendix. If you answer \answerYes{} to a question, in the justification please point to the section(s) where related material for the question can be found.

IMPORTANT, please:
\begin{itemize}
    \item {\bf Delete this instruction block, but keep the section heading ``NeurIPS Paper Checklist"},
    \item  {\bf Keep the checklist subsection headings, questions/answers and guidelines below.}
    \item {\bf Do not modify the questions and only use the provided macros for your answers}.
\end{itemize}


\begin{enumerate}

\item {\bf Claims}
    \item[] Question: Do the main claims made in the abstract and introduction accurately reflect the paper's contributions and scope?
    \item[] Answer:  \answerYes{} 
    \item[] Justification:  \answerYes{}
    \item[] Guidelines:
    \begin{itemize}
        \item The answer NA means that the abstract and introduction do not include the claims made in the paper.
        \item The abstract and/or introduction should clearly state the claims made, including the contributions made in the paper and important assumptions and limitations. A No or NA answer to this question will not be perceived well by the reviewers. 
        \item The claims made should match theoretical and experimental results, and reflect how much the results can be expected to generalize to other settings. 
        \item It is fine to include aspirational goals as motivation as long as it is clear that these goals are not attained by the paper. 
    \end{itemize}

\item {\bf Limitations}
    \item[] Question: Does the paper discuss the limitations of the work performed by the authors?
    \item[] Answer: \answerYes{} 
    \item[] Justification: \answerYes{}
    \item[] Guidelines:
    \begin{itemize}
        \item The answer NA means that the paper has no limitation while the answer No means that the paper has limitations, but those are not discussed in the paper. 
        \item The authors are encouraged to create a separate "Limitations" section in their paper.
        \item The paper should point out any strong assumptions and how robust the results are to violations of these assumptions (e.g., independence assumptions, noiseless settings, model well-specification, asymptotic approximations only holding locally). The authors should reflect on how these assumptions might be violated in practice and what the implications would be.
        \item The authors should reflect on the scope of the claims made, e.g., if the approach was only tested on a few datasets or with a few runs. In general, empirical results often depend on implicit assumptions, which should be articulated.
        \item The authors should reflect on the factors that influence the performance of the approach. For example, a facial recognition algorithm may perform poorly when image resolution is low or images are taken in low lighting. Or a speech-to-text system might not be used reliably to provide closed captions for online lectures because it fails to handle technical jargon.
        \item The authors should discuss the computational efficiency of the proposed algorithms and how they scale with dataset size.
        \item If applicable, the authors should discuss possible limitations of their approach to address problems of privacy and fairness.
        \item While the authors might fear that complete honesty about limitations might be used by reviewers as grounds for rejection, a worse outcome might be that reviewers discover limitations that aren't acknowledged in the paper. The authors should use their best judgment and recognize that individual actions in favor of transparency play an important role in developing norms that preserve the integrity of the community. Reviewers will be specifically instructed to not penalize honesty concerning limitations.
    \end{itemize}

\item {\bf Theory assumptions and proofs}
    \item[] Question: For each theoretical result, does the paper provide the full set of assumptions and a complete (and correct) proof?
    \item[] Answer: \answerNA{} 
    \item[] Justification: \answerNA{}
    \item[] Guidelines:
    \begin{itemize}
        \item The answer NA means that the paper does not include theoretical results. 
        \item All the theorems, formulas, and proofs in the paper should be numbered and cross-referenced.
        \item All assumptions should be clearly stated or referenced in the statement of any theorems.
        \item The proofs can either appear in the main paper or the supplemental material, but if they appear in the supplemental material, the authors are encouraged to provide a short proof sketch to provide intuition. 
        \item Inversely, any informal proof provided in the core of the paper should be complemented by formal proofs provided in appendix or supplemental material.
        \item Theorems and Lemmas that the proof relies upon should be properly referenced. 
    \end{itemize}

    \item {\bf Experimental result reproducibility}
    \item[] Question: Does the paper fully disclose all the information needed to reproduce the main experimental results of the paper to the extent that it affects the main claims and/or conclusions of the paper (regardless of whether the code and data are provided or not)?
    \item[] Answer: \answerYes{} 
    \item[] Justification:\answerYes{}
    \item[] Guidelines:
    \begin{itemize}
        \item The answer NA means that the paper does not include experiments.
        \item If the paper includes experiments, a No answer to this question will not be perceived well by the reviewers: Making the paper reproducible is important, regardless of whether the code and data are provided or not.
        \item If the contribution is a dataset and/or model, the authors should describe the steps taken to make their results reproducible or verifiable. 
        \item Depending on the contribution, reproducibility can be accomplished in various ways. For example, if the contribution is a novel architecture, describing the architecture fully might suffice, or if the contribution is a specific model and empirical evaluation, it may be necessary to either make it possible for others to replicate the model with the same dataset, or provide access to the model. In general. releasing code and data is often one good way to accomplish this, but reproducibility can also be provided via detailed instructions for how to replicate the results, access to a hosted model (e.g., in the case of a large language model), releasing of a model checkpoint, or other means that are appropriate to the research performed.
        \item While NeurIPS does not require releasing code, the conference does require all submissions to provide some reasonable avenue for reproducibility, which may depend on the nature of the contribution. For example
        \begin{enumerate}
            \item If the contribution is primarily a new algorithm, the paper should make it clear how to reproduce that algorithm.
            \item If the contribution is primarily a new model architecture, the paper should describe the architecture clearly and fully.
            \item If the contribution is a new model (e.g., a large language model), then there should either be a way to access this model for reproducing the results or a way to reproduce the model (e.g., with an open-source dataset or instructions for how to construct the dataset).
            \item We recognize that reproducibility may be tricky in some cases, in which case authors are welcome to describe the particular way they provide for reproducibility. In the case of closed-source models, it may be that access to the model is limited in some way (e.g., to registered users), but it should be possible for other researchers to have some path to reproducing or verifying the results.
        \end{enumerate}
    \end{itemize}

\item {\bf Open access to data and code}
    \item[] Question: Does the paper provide open access to the data and code, with sufficient instructions to faithfully reproduce the main experimental results, as described in supplemental material?
    \item[] Answer: \answerNA{}
    \item[] Justification: \answerNA{}
    \item[] Guidelines:
    \begin{itemize}
        \item The answer NA means that paper does not include experiments requiring code.
        \item Please see the NeurIPS code and data submission guidelines (\url{https://nips.cc/public/guides/CodeSubmissionPolicy}) for more details.
        \item While we encourage the release of code and data, we understand that this might not be possible, so “No” is an acceptable answer. Papers cannot be rejected simply for not including code, unless this is central to the contribution (e.g., for a new open-source benchmark).
        \item The instructions should contain the exact command and environment needed to run to reproduce the results. See the NeurIPS code and data submission guidelines (\url{https://nips.cc/public/guides/CodeSubmissionPolicy}) for more details.
        \item The authors should provide instructions on data access and preparation, including how to access the raw data, preprocessed data, intermediate data, and generated data, etc.
        \item The authors should provide scripts to reproduce all experimental results for the new proposed method and baselines. If only a subset of experiments are reproducible, they should state which ones are omitted from the script and why.
        \item At submission time, to preserve anonymity, the authors should release anonymized versions (if applicable).
        \item Providing as much information as possible in supplemental material (appended to the paper) is recommended, but including URLs to data and code is permitted.
    \end{itemize}

\item {\bf Experimental setting/details}
    \item[] Question: Does the paper specify all the training and test details (e.g., data splits, hyperparameters, how they were chosen, type of optimizer, etc.) necessary to understand the results?
    \item[] Answer: \answerYes{} 
    \item[] Justification: \answerYes{}
    \item[] Guidelines:
    \begin{itemize}
        \item The answer NA means that the paper does not include experiments.
        \item The experimental setting should be presented in the core of the paper to a level of detail that is necessary to appreciate the results and make sense of them.
        \item The full details can be provided either with the code, in appendix, or as supplemental material.
    \end{itemize}

\item {\bf Experiment statistical significance}
    \item[] Question: Does the paper report error bars suitably and correctly defined or other appropriate information about the statistical significance of the experiments?
    \item[] Answer: \answerYes{} 
    \item[] Justification: \answerYes{}
    \item[] Guidelines:
    \begin{itemize}
        \item The answer NA means that the paper does not include experiments.
        \item The authors should answer "Yes" if the results are accompanied by error bars, confidence intervals, or statistical significance tests, at least for the experiments that support the main claims of the paper.
        \item The factors of variability that the error bars are capturing should be clearly stated (for example, train/test split, initialization, random drawing of some parameter, or overall run with given experimental conditions).
        \item The method for calculating the error bars should be explained (closed form formula, call to a library function, bootstrap, etc.)
        \item The assumptions made should be given (e.g., Normally distributed errors).
        \item It should be clear whether the error bar is the standard deviation or the standard error of the mean.
        \item It is OK to report 1-sigma error bars, but one should state it. The authors should preferably report a 2-sigma error bar than state that they have a 96\% CI, if the hypothesis of Normality of errors is not verified.
        \item For asymmetric distributions, the authors should be careful not to show in tables or figures symmetric error bars that would yield results that are out of range (e.g. negative error rates).
        \item If error bars are reported in tables or plots, The authors should explain in the text how they were calculated and reference the corresponding figures or tables in the text.
    \end{itemize}

\item {\bf Experiments compute resources}
    \item[] Question: For each experiment, does the paper provide sufficient information on the computer resources (type of compute workers, memory, time of execution) needed to reproduce the experiments?
    \item[] Answer: \answerYes{} 
    \item[] Justification:\answerYes{}
    \item[] Guidelines:
    \begin{itemize}
        \item The answer NA means that the paper does not include experiments.
        \item The paper should indicate the type of compute workers CPU or GPU, internal cluster, or cloud provider, including relevant memory and storage.
        \item The paper should provide the amount of compute required for each of the individual experimental runs as well as estimate the total compute. 
        \item The paper should disclose whether the full research project required more compute than the experiments reported in the paper (e.g., preliminary or failed experiments that didn't make it into the paper). 
    \end{itemize}
    
\item {\bf Code of ethics}
    \item[] Question: Does the research conducted in the paper conform, in every respect, with the NeurIPS Code of Ethics \url{https://neurips.cc/public/EthicsGuidelines}?
    \item[] Answer: \answerYes{} 
    \item[] Justification: \answerYes{}
    \item[] Guidelines:
    \begin{itemize}
        \item The answer NA means that the authors have not reviewed the NeurIPS Code of Ethics.
        \item If the authors answer No, they should explain the special circumstances that require a deviation from the Code of Ethics.
        \item The authors should make sure to preserve anonymity (e.g., if there is a special consideration due to laws or regulations in their jurisdiction).
    \end{itemize}

\item {\bf Broader impacts}
    \item[] Question: Does the paper discuss both potential positive societal impacts and negative societal impacts of the work performed?
    \item[] Answer: \answerYes{} 
    \item[] Justification:  \answerYes{}
    \item[] Guidelines:
    \begin{itemize}
        \item The answer NA means that there is no societal impact of the work performed.
        \item If the authors answer NA or No, they should explain why their work has no societal impact or why the paper does not address societal impact.
        \item Examples of negative societal impacts include potential malicious or unintended uses (e.g., disinformation, generating fake profiles, surveillance), fairness considerations (e.g., deployment of technologies that could make decisions that unfairly impact specific groups), privacy considerations, and security considerations.
        \item The conference expects that many papers will be foundational research and not tied to particular applications, let alone deployments. However, if there is a direct path to any negative applications, the authors should point it out. For example, it is legitimate to point out that an improvement in the quality of generative models could be used to generate deepfakes for disinformation. On the other hand, it is not needed to point out that a generic algorithm for optimizing neural networks could enable people to train models that generate Deepfakes faster.
        \item The authors should consider possible harms that could arise when the technology is being used as intended and functioning correctly, harms that could arise when the technology is being used as intended but gives incorrect results, and harms following from (intentional or unintentional) misuse of the technology.
        \item If there are negative societal impacts, the authors could also discuss possible mitigation strategies (e.g., gated release of models, providing defenses in addition to attacks, mechanisms for monitoring misuse, mechanisms to monitor how a system learns from feedback over time, improving the efficiency and accessibility of ML).
    \end{itemize}
    
\item {\bf Safeguards}
    \item[] Question: Does the paper describe safeguards that have been put in place for responsible release of data or models that have a high risk for misuse (e.g., pretrained language models, image generators, or scraped datasets)?
    \item[] Answer: \answerNA{} 
    \item[] Justification: \answerNA{}
    \item[] Guidelines:
    \begin{itemize}
        \item The answer NA means that the paper poses no such risks.
        \item Released models that have a high risk for misuse or dual-use should be released with necessary safeguards to allow for controlled use of the model, for example by requiring that users adhere to usage guidelines or restrictions to access the model or implementing safety filters. 
        \item Datasets that have been scraped from the Internet could pose safety risks. The authors should describe how they avoided releasing unsafe images.
        \item We recognize that providing effective safeguards is challenging, and many papers do not require this, but we encourage authors to take this into account and make a best faith effort.
    \end{itemize}

\item {\bf Licenses for existing assets}
    \item[] Question: Are the creators or original owners of assets (e.g., code, data, models), used in the paper, properly credited and are the license and terms of use explicitly mentioned and properly respected?
    \item[] Answer: \answerNA{} 
    \item[] Justification: \answerNA{}
    \item[] Guidelines:
    \begin{itemize}
        \item The answer NA means that the paper does not use existing assets.
        \item The authors should cite the original paper that produced the code package or dataset.
        \item The authors should state which version of the asset is used and, if possible, include a URL.
        \item The name of the license (e.g., CC-BY 4.0) should be included for each asset.
        \item For scraped data from a particular source (e.g., website), the copyright and terms of service of that source should be provided.
        \item If assets are released, the license, copyright information, and terms of use in the package should be provided. For popular datasets, \url{paperswithcode.com/datasets} has curated licenses for some datasets. Their licensing guide can help determine the license of a dataset.
        \item For existing datasets that are re-packaged, both the original license and the license of the derived asset (if it has changed) should be provided.
        \item If this information is not available online, the authors are encouraged to reach out to the asset's creators.
    \end{itemize}

\item {\bf New assets}
    \item[] Question: Are new assets introduced in the paper well documented and is the documentation provided alongside the assets?
    \item[] Answer: \answerNA{} 
    \item[] Justification: \answerNA{}
    \item[] Guidelines:
    \begin{itemize}
        \item The answer NA means that the paper does not release new assets.
        \item Researchers should communicate the details of the dataset/code/model as part of their submissions via structured templates. This includes details about training, license, limitations, etc. 
        \item The paper should discuss whether and how consent was obtained from people whose asset is used.
        \item At submission time, remember to anonymize your assets (if applicable). You can either create an anonymized URL or include an anonymized zip file.
    \end{itemize}

\item {\bf Crowdsourcing and research with human subjects}
    \item[] Question: For crowdsourcing experiments and research with human subjects, does the paper include the full text of instructions given to participants and screenshots, if applicable, as well as details about compensation (if any)? 
    \item[] Answer: \answerNA{} 
    \item[] Justification: \answerNA{}
    \item[] Guidelines:
    \begin{itemize}
        \item The answer NA means that the paper does not involve crowdsourcing nor research with human subjects.
        \item Including this information in the supplemental material is fine, but if the main contribution of the paper involves human subjects, then as much detail as possible should be included in the main paper. 
        \item According to the NeurIPS Code of Ethics, workers involved in data collection, curation, or other labor should be paid at least the minimum wage in the country of the data collector. 
    \end{itemize}

\item {\bf Institutional review board (IRB) approvals or equivalent for research with human subjects}
    \item[] Question: Does the paper describe potential risks incurred by study participants, whether such risks were disclosed to the subjects, and whether Institutional Review Board (IRB) approvals (or an equivalent approval/review based on the requirements of your country or institution) were obtained?
    \item[] Answer: \answerNA{} 
    \item[] Justification: \answerNA{}
    \item[] Guidelines:
    \begin{itemize}
        \item The answer NA means that the paper does not involve crowdsourcing nor research with human subjects.
        \item Depending on the country in which research is conducted, IRB approval (or equivalent) may be required for any human subjects research. If you obtained IRB approval, you should clearly state this in the paper. 
        \item We recognize that the procedures for this may vary significantly between institutions and locations, and we expect authors to adhere to the NeurIPS Code of Ethics and the guidelines for their institution. 
        \item For initial submissions, do not include any information that would break anonymity (if applicable), such as the institution conducting the review.
    \end{itemize}

\item {\bf Declaration of LLM usage}
    \item[] Question: Does the paper describe the usage of LLMs if it is an important, original, or non-standard component of the core methods in this research? Note that if the LLM is used only for writing, editing, or formatting purposes and does not impact the core methodology, scientific rigorousness, or originality of the research, declaration is not required.
    \item[] Answer: \answerNA{} 
    \item[] Justification: \answerNA{}
    \item[] Guidelines:
    \begin{itemize}
        \item The answer NA means that the core method development in this research does not involve LLMs as any important, original, or non-standard components.
        \item Please refer to our LLM policy (\url{https://neurips.cc/Conferences/2025/LLM}) for what should or should not be described.
    \end{itemize}

\end{enumerate}

\end{document}

%% file: section/01_intro.tex
\section{Introduction}
Recent studies suggest that LLMs possess strong structural reasoning abilities~\cite{nlgraph,talklikeagraph}. To investigate the underlying reasons of this phenomenon, existing work has provided both theoretical and empirical insights within 
simplified graph reasoning settings~\cite{transformer_struggle,wang2024alpine,dai2025sequence}, where the reasoning process is studied using decoder-only Transformers trained from scratch on graphs represented solely by node IDs. However, existing analyses remain case-specific, as different graph reasoning tasks often require distinct analytical methods. These studies still do not provide a unified understanding of the underlying mechanisms by which models perform reasoning over explicit graph structures. Therefore, understanding how decoder-only Transformers solve graph reasoning tasks requires continued investigation into their unified mechanisms.

Specifically, for graph reasoning tasks, the mechanisms underlying 
transformers' performance in path reasoning and pattern extraction have been attributed to edge memorization~\cite{wang2024alpine} and progressive filtration across layers~\cite{dai2025sequence}. Since both tasks demonstrate that decoder-only transformers possess a fundamental capacity for structural understanding, it is reasonable to expect that a consistent underlying mechanism governs their reasoning over explicit graph-like structures encoded in textual sequences. This motivates the need for a unified interpretability framework capable of revealing shared mechanisms across diverse graph reasoning tasks.
Concurrently, the circuit-tracing interpretation framework~\cite{ameisen2025circuit} has been successfully applied to various language tasks 
. This framework analyzes how information flows through a Transformer by identifying the specific neurons and layer interactions that contribute to a model’s predictions, uncovering latent reasoning structures that contribute to model performance.
Due to the inherent irregularity of language tokens, examining the relationship between implicit structures and explicit structural representations offers valuable insight into how transformers process structured information.

Building on these insights, we apply the circuit-tracing interpretation framework to graph reasoning tasks to investigate whether decoder-only Transformers exhibit consistent interpretability across such tasks.
Specifically, we conduct case studies using circuit-tracer analyses to examine how these models reason over explicit structures in tasks such as path reasoning, pattern extraction, and attributed graph reasoning.
Through visualization-based analyses, we identify layer-wise \textbf{token merging} and  \textbf{structure memorization} as the primary mechanisms underlying the models’ reasoning behavior. Specifically, token merging indicates that Transformers progressively combine tokens to construct substructures relevant to the prediction task. In parallel, structure memorization reveals that the model’s predictions also rely on patterns learned from the training data, suggesting a form of retrieval or recall from previously seen structures.
To further support this, we perform quantitative analyses to measure the prevalence of these phenomena across different graph reasoning tasks, considering variations in graph density and the hidden dimension size of the underlying Transformer backbone. In summary, our contributions are as follows:

1. We apply circuit-tracing methods to graph reasoning tasks to obtain a unified interpretation of how and why decoder-only Transformers are able to solve them.

2. We provide both visualization and statistical analyses to interpret the underlying mechanisms of token merging and structural memorization for decoder-only transformers in graph reasoning tasks.

3.  We further analyze the effects of graph density on token merging and the influence of Transformer hidden size on structural memorization.

%% file: section/03_Experiments.tex
\input{table/task_table}

\section{Experiments}
\subsection{Experiment setting}

First, following previous work~\cite{wang2024alpine,dai2025sequence}, we generate synthetic graph datasets to train the transformers. Specifically, we construct a graph $G$ with $|N|$ nodes. Then, we sample the subgraphs $G'$ from the generated graph to construct the training samples, each containing $|N'|<|N|$ nodes. We focus on three fundamental graph reasoning tasks: path reasoning, attributed graph reasoning, and pattern extraction. For each training sample, the input is formatted as a sequence of the form "<Graph(EL)><Question><Answer>" following the definition in~\citep{dai2025sequence}
where the question and answer components vary depending on the task. Examples of task-specific prompts are provided in Table~\ref{tab:graph_samples}. We adopt the GPT-2 model and train transcoders across multiple layers according to~\citep{ameisen2025circuit}. The data and training details are in Appendix~\ref{app:exp}.




\vspace{-0.1cm}

\subsection{Visualized tracers in graph reasoning tasks}
\begin{figure}[htbp]
\vspace{-0.2cm}
  \centering

  \begin{minipage}[c]{0.45\textwidth}
   \centering
    
    \includegraphics[width=\linewidth]{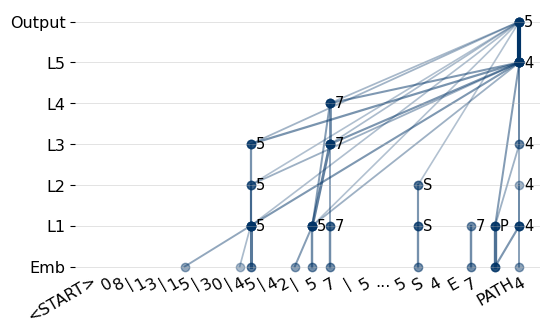}
    \caption{Circuit tracer in the path reasoning task. ``L'' denotes the layers. The predicted path is 4 → 5 → 7, with the model currently predicting token 5.} 
    \label{fig:path_vis}

  \end{minipage}
\hfill
  \hfill
  \begin{minipage}[c]{0.45\textwidth}
    \centering
    \includegraphics[width=\linewidth]{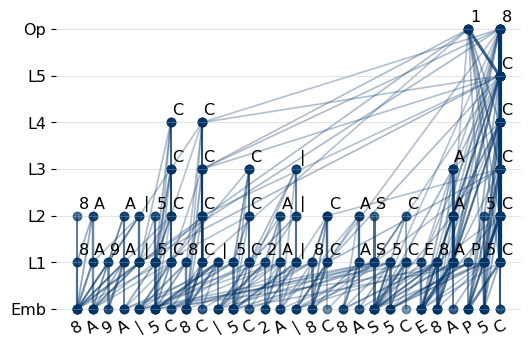} \\
    \caption{Circuit tracer in the attributed graph reasoning task. The predicted path is 5C → 8C → 8A, with the current token being 8 in the token 8C.}
    \label{fig:attr_vis}
  \end{minipage}

\end{figure}

To reveal the internal reasoning structures of Transformers, we first visualize circuit traces across three representative graph reasoning tasks—\textbf{path reasoning}, \textbf{attributed graph reasoning}, and \textbf{pattern extraction}, as shown in Figure~\ref{fig:path_vis}, Figure~\ref{fig:attr_vis}, and Appendix~\ref{app:figs} Figure~\ref{fig:patt_vis}, respectively.

\begin{figure}[htbp]
\vspace{-0.2cm}
  \centering

  \begin{minipage}[c]{0.25\textwidth}
    \centering
    \includegraphics[width=\linewidth]{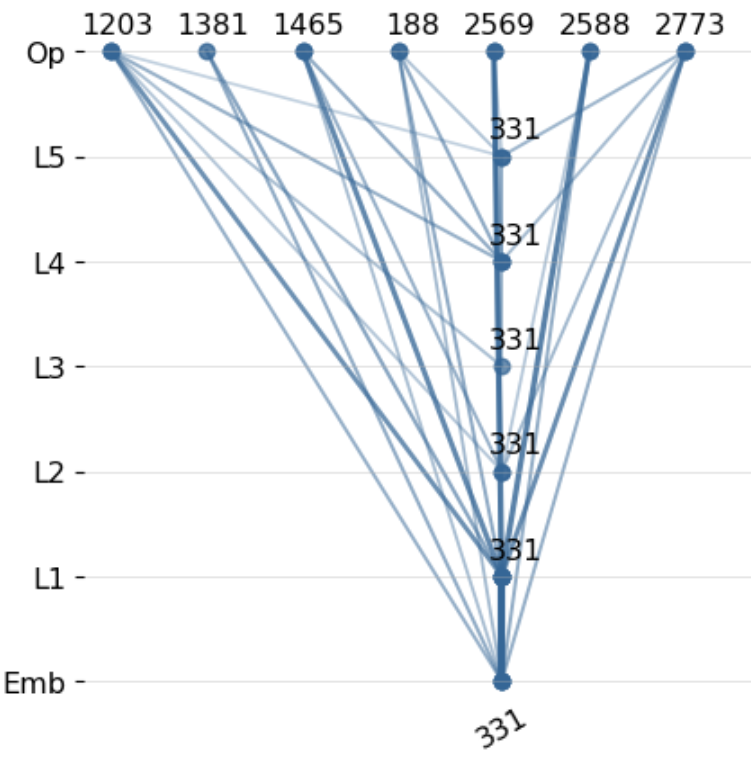} \\
    \caption{Circuit tracer reveals structural memorization: different layers store information about the 1-hop neighbors of node 331.}
    \label{fig:mem_vis}
  \end{minipage}
\hfill
\begin{minipage}[c]{0.45\textwidth}
   \centering
    
    \includegraphics[width=\linewidth]{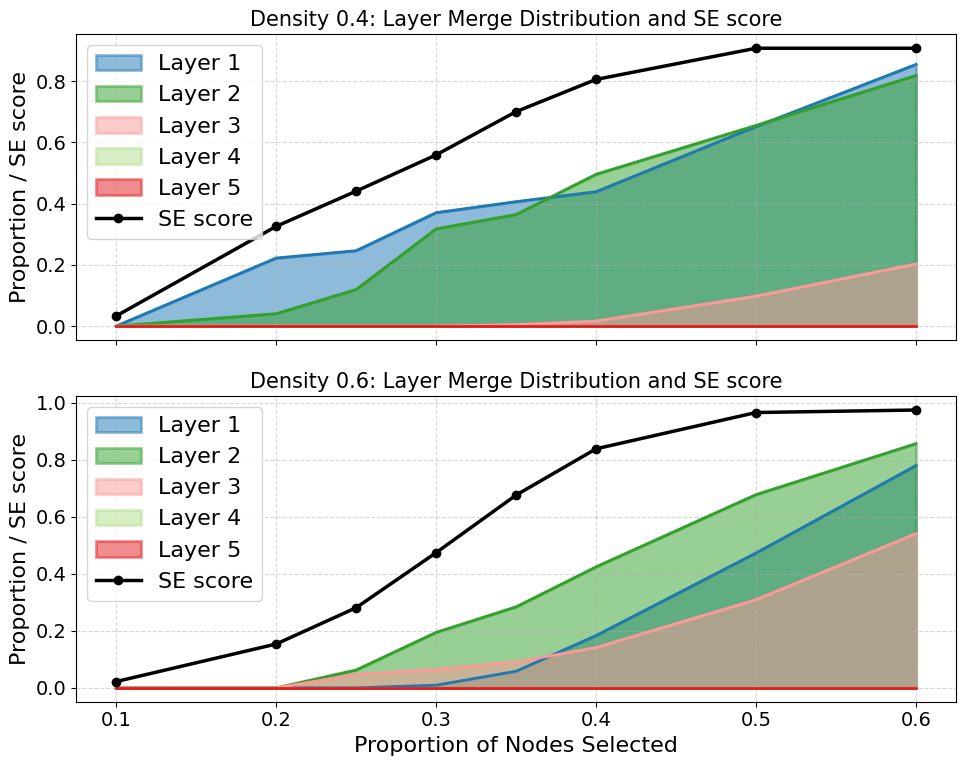}
    \caption{Edge gathering in path reasoning task. The evaluations are on graphs with 0.4 and 0.6, respectively.}
    \label{fig:path_reasoning}
  \end{minipage}
  \hfill
    \begin{minipage}[c]{0.25\textwidth}
    \centering
    \includegraphics[width=\linewidth]{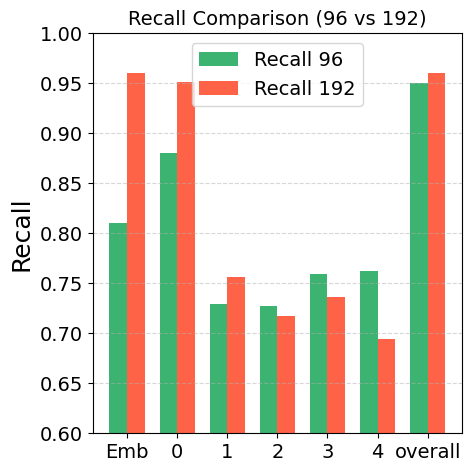} \\
    \caption{The Recall of structure memorization under different hidden dimensions}
    \label{fig:rec}
  \end{minipage}
  \vspace{-0.6cm}
\end{figure}

The visualizations reveal token merging as a core mechanism in the reasoning process across various graph tasks. For example, in the path reasoning task (Figure~\ref{fig:path_vis}), given a graph and a query specifying start node 4 and end node 7, the expected path is 4 → 5 → 7. The model is tasked with predicting the next token node 5. Circuit tracing shows that the Transformer merges the edges (4, 5) and (5, 7), enabling it to correctly identify node 5 as the next step toward the target. Similarly, in the attributed path reasoning task shown in Figure~\ref{fig:attr_vis}, we observe that the relevant edges are also merged, with their associated attributes included in the merged representations.
For instance, the target edge (5C, 8C) is clearly highlighted, indicating its role in guiding the model's prediction. 
In the pattern extraction task shown in Appendix~\ref{app:figs} Figure~\ref{fig:patt_vis}, token merging is also observed at higher layers. For example, when the predicted pattern is (27, 23, 40), the tokens corresponding to nodes 23 and 40 are merged into token 27, a process that becomes more prominent in the upper layers of the model.

In addition to token merging, we observe that Transformers progressively memorize structural information during training them to solve path reasoning tasks, as detailed in Appendix~\ref{app:mem_train}. After training on path reasoning tasks, we prompt the Transformer with a single node ID and assess whether it can predict the next token, modeling the probability of its 1-hop neighbors. To further interpret this memorization, we analyze the circuit tracer results and examine how multiple layers contribute to preserving neighborhood information, as visualized in Figure~\ref{fig:mem_vis}. In the large-graph path reasoning setting, Transformers are capable of recalling neighbor nodes when the prompt is a central node. Importantly, different layers contribute differently to this memorization. For example, node 1381 is already recalled in the embedding and first layers.
while node 1203 requires additional support from every layer. This layer-wise contribution reflects how structural information is distributed and retrieved during inference.

\vspace{-0.1cm}
\subsection{Quantified evaluations}

From the visualizations, we observe that the circuit tracers reveal the reasoning process on a case-by-case basis. In this section, we present quantitative evaluations to demonstrate that token merging and structure memorization are consistent behaviors at the dataset level. Furthermore, we analyze how these two mechanisms are affected by varying graph densities and model sizes.



    



\vspace{-0.2cm}
\paragraph{Token Merging} 
The token merging mechanism consistently emerges across various graph reasoning tasks, serving to summarize substructures relevant to the model’s predictions, such as edge pairs in path reasoning and candidate patterns in substructure extraction tasks. To evaluate this behavior, we measure the alignment between the selected and expected tokens using the metric $\mathrm{S_E} = \frac{N_{\text{select}}}{N_{\text{pred}}}$, where $N_{\text{pred}}$ denotes the number of the tokens are extracted by the circuit tracer, and $N_{\text{select}}$ represents the number of the tokens that exactly can provide the evidences for the predictions. For example, these include triangle patterns in pattern extraction, attributed nodes and edges merged in attributed graph reasoning, and edges gathered in the path reasoning task. The corresponding results are shown in Appendix~\ref{app:figs} Appendix~\ref{app:figs} Figure~\ref{fig:substructure}, and Table~\ref{tab:attr_graph}, Appendix~\ref{app:figs} Figure~\ref{fig:path_reasoning}, respectively, where the layers at which token merging occurs vary across tasks and token types.


Appendix~\ref{app:figs} Figure~\ref{fig:substructure} illustrates the pattern extraction task. We observe that token merging occurs progressively across layers, with multiple merging operations taking place in the shallow layers. During this stage, the number of included nodes increases significantly, but the SE score remains stable as the selected node ratio increases from 0.7 to 0.9.
In attributed graph reasoning, shown in Appendix~\ref{app:figs} Table~\ref{tab:attr_graph}, tokens corresponding to start nodes, end nodes, and edge attributes are typically merged around layer 3. Meanwhile, path reasoning tasks in Figure~\ref{fig:path_reasoning} reveal a density-dependent pattern: for graphs with low degree (e.g., density 0.4), many relevant edges are detected in early layers (layer 1 or 2), whereas in denser graphs (e.g., density 0.6), the model relies more heavily on deeper processing, particularly at layer 3. In conclusion, token merging is consistently observed across various tasks, and as graph density increases, the merging tends to occur at higher layers of the Transformer.

\vspace{-0.1cm}
\paragraph{Structure Memorization} 
The structure memorization mechanism suggests that next-token predictions are influenced by structural patterns learned from the training data. We observe that Transformers are capable of memorizing graph structures, as evidenced by their ability to recall potential neighbor nodes during inference. Therefore, we apply Precision and Recall to discuss whether the transformers memorize the correct neighbors and cover all of the neighbors, respectively.
Notably, the memorization spans multiple layers, indicating that structural information is preserved and propagated throughout the model rather than being localized to a specific layer.
To evaluate this behavior, we assess whether the neighbors predicted by the model correspond to those in the original graph. Given that hidden size is often a critical factor in neural network memorization capacity, we compare models with different hidden dimensions, specifically 96 and 192, as shown in Appendix~\ref{app:figs} Figure~\ref{fig:pre} and Figure~\ref{fig:rec}.
In both settings, the precision is consistently high, indicating that Transformers can accurately memorize 1-hop neighbors across multiple layers. However, the recall patterns differ. Although the overall recall scores are similar, the distribution of memorized neighbors across layers varies. In models with smaller hidden dimensions, memorization tends to be more evenly distributed across layers. In contrast, models with larger hidden dimensions exhibit strong memorization even in shallow layers; notably, the embedding layer alone is capable of capturing all 1-hop neighbors. This suggests that higher-dimensional embeddings are sufficient to encode local structural information without requiring deeper processing. In summary, Transformers are capable of memorizing graph structures, and larger hidden sizes encourage this neighbor memorization to be concentrated in the lower layers of the model.



%% file: table/task_table.tex
\begin{table}[]
\vspace{-1.5cm}
\caption{Examples of graph samples in textual format across different tasks }
\label{tab:graph_samples}
\resizebox{\linewidth}{!}{%
\begin{tabular}{llll}
\toprule
                                 & Path Reasoning              & Attributed Graph Reasoning                  & Substructure Extraction     \\
\midrule
Example &{\raisebox{-0.7cm}{\includegraphics[width=0.15\textwidth]{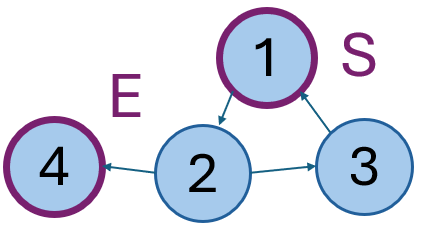}}} &{\raisebox{-0.7cm}{\includegraphics[width=0.15\textwidth]{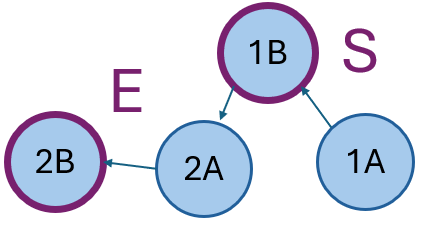}}} &{\raisebox{-0.7cm}{\includegraphics[width=0.15\textwidth]{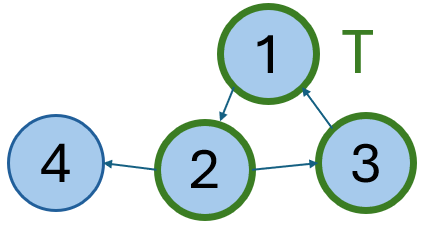}}} \\
Graph\textless{}EL\textgreater{} & Node index list: 1 2|2 3|3 1|2 4   & Node index with attributes: 1A 1B|1B 2A|2A 2B & Node index list: 1 2|2 3|3 1|2 4   \\
Prompt                           & Start and end nodes: S 1, E 4 & Start and end nodes: S 1B, E 2B               & Substructure (Triangle): T   \\
Answer                           & Nodes in the shortest path: 1 2 4         & Nodes in the shortest path: 1B 2A 2B                      & Nodes in substructure: 1 2 3 \\
\midrule
Description                      & \makecell[l]{Given a graph and \\ target nodes, predict the path} 
                                 & \makecell[l]{Given an attributed graph and \\ target nodes, predict the path} 
                                 & \makecell[l]{Given a graph and a substructure symbol, \\ predict all matching substructures} \\
\bottomrule
\end{tabular}%
}
\vspace{-0.5cm}
\end{table}

%% file: section/02_Related_work_and_Background.tex
\vspace{-0.2cm}
\section{Related work and Background}
\vspace{-0.1cm}
\paragraph{Mechanism of graph structure understanding} Recent studies suggest that large language models (LLMs) possess the capability to understand graph structures~\citep{nlgraph,graphPatt}. To investigate the underlying reasons for this ability, current research often focuses on simplified graph data and decoder-only transformers, aiming to uncover the mechanisms behind such capabilities~\citep{sanford2024understanding,bachmann2024pitfalls}. For instance,~\citep{wang2024alpine,transformer_struggle} suggest that LLMs perform path reasoning by effectively searching for relevant edges, while~\citep{dai2025sequence} argues that LLMs can extract substructures from the input. Despite these varying perspectives, it remains unclear whether there exists a unified framework for understanding the mechanisms by which LLMs process graph structures.
\vspace{-0.1cm}
\paragraph{Interpretation of LLMs}

Understanding the internal mechanisms of Transformer models has long been a focus of interpretability research~\citep{vig2019analyzing,zhao2024explainability}. Early interpretability studies on BERT revealed that different layers capture distinct linguistic properties: lower layers attend to local syntax, while higher layers progressively aggregate global semantics~\citep{clark2019does, tenney2019bert}. Probing classifiers and attention-based analyses further demonstrated how information is organized hierarchically across depth~\citep{rogers2020primer}. In decoder-only Transformers, this line of work evolved into circuit tracing methods such as Transcoder~\citep{anthropic2023transcoder,dunefsky2024transcoders}, which recover token-level causal paths to explain autoregressive reasoning. Building on this, recent approaches introduce attributed graphs to represent token interactions across layers, capturing both semantic roles and attention dynamics~\citep{ameisen2025circuit}. Our work extends this direction by constructing attributed graphs from circuit traces to interpret graph reasoning tasks, revealing token merging and structure memorization dynamics in graph reasoning tasks.
\vspace{-0.2cm}

%% file: table/parameters.tex
\begin{table}[]
\caption{Summarization of Experiment settings}
\label{tab:exp_set}
\resizebox{\linewidth}{!}{\begin{tabular}{ll|llll}
\toprule
\multirow{2}{*}{Module}              & \multirow{2}{*}{Parameter}           & \multicolumn{2}{l}{Path reasoning}                   & \multicolumn{1}{l}{\multirow{2}{*}{Attributed  reasoning}} & \multicolumn{1}{l}{\multirow{2}{*}{Substructure extraction}} \\

                               &                             & \multicolumn{1}{l}{Tiny} & \multicolumn{1}{l}{Large} & \multicolumn{1}{l}{}                                       & \multicolumn{1}{l}{}                                         \\\midrule
\multirow{3}{*}{Graph}         & Density                     & 0.4                      & 0.002                     & 0.2                                                        & 0.1                                                          \\
                               & Node num                   & 50                       & 3000                      & 30                                                         & 100                                                          \\
                               & Max node number in subgraph & 10                       & 10                        & 5                                                          & 5                                                            \\
\multirow{3}{*}{Transformer}   & Hidden size                 & 96                       & 96                        & 96                                                         & 96                                                           \\
                               & Max length                  & 256                      & 96                        & 96                                                         & 96                                                           \\
                               & Basic Acc                   & 0.99                     & 0.96                      & 0.94                                                       & 0.92                                                         \\
\multirow{3}{*}{Transcoder}    & L1 coefficent              & 0.0005                   & 0.0005                    & 0.0005                                                     & 0.0005                                                       \\
                               & Dead nueron num             & 50                       & 50                        & 50                                                         & 50                                                           \\
                               & Hidden size                 & 192                      & 192                       & 192                                                        & 192                                                          \\
\multirow{3}{*}{Vis parameter} & Node threshold                   & 0.86                     & 0.8                       & 0.8                                                        & 0.8                                                          \\
                               & Edge ratio                  & 0.48                     & 0.9                       & 0.99                                                       & 0.9                                                          \\
                               & Edge threshold                   & 0.1                      & 0.2                       & 0.1                                                        & 0.4                                                         
\\
\bottomrule
\end{tabular}}
\end{table}